\documentclass{article}
\PassOptionsToPackage{numbers, compress}{natbib}

\usepackage{graphicx}
\usepackage{booktabs}
\newcommand{\myparagraph}[1]{\vspace{0.0em}\noindent\vspace{0.0em}\textbf{#1}}
\usepackage[accsupp]{axessibility}  
\usepackage{multirow}
\usepackage{tabularx}

\usepackage{enumitem}
\usepackage[pagebackref,breaklinks,colorlinks]{hyperref}
\usepackage{url}
\usepackage{orcidlink}

\usepackage[preprint]{neurips_2024}

\usepackage[utf8]{inputenc} 
\usepackage[T1]{fontenc}    
\usepackage{hyperref}       
\usepackage{url}            
\usepackage{booktabs}       
\usepackage{amsfonts}       
\usepackage{nicefrac}       
\usepackage{microtype}      
\usepackage{xcolor}         
\usepackage{wrapfig,lipsum,booktabs}

\usepackage{enumitem}
\usepackage[pagebackref,breaklinks,colorlinks]{hyperref}
\usepackage{url}
\usepackage{orcidlink}
\usepackage{multirow}
\usepackage{tabularx}
\usepackage{pifont}
\usepackage[preprint]{neurips_2024}

\usepackage[utf8]{inputenc} 
\usepackage[T1]{fontenc}    
\usepackage{hyperref}       
\usepackage{url}            
\usepackage{booktabs}       
\usepackage{amsfonts}       
\usepackage{nicefrac}       
\usepackage{microtype}      
\usepackage{xcolor}         
\usepackage{wrapfig,lipsum,booktabs}

\usepackage{url}
\usepackage{caption}

\title{Inv-Adapter: ID Customization Generation via Image Inversion and Lightweight Adapter}

\author{%
  Peng Xing  \\
  Nanjing University of Science and Technology \\
  \texttt{xingp\_ng@njust.edu.cn} \\
  \And
  Ning Wang \\
  Huawei Technologies Ltd. \\
  \texttt{wn6149@mail.ustc.edu.cn} \\
  \And
  JianBo Ouyang\\
  Huawei Technologies Ltd.\\
  \texttt{ouyjb@mail.ustc.edu.cn} \\
  \And
  Zechao Li\\
  Nanjing University of Science and Technology\\
  \texttt{zechao.li@njust.edu.cn}\\
}

\begin{document}

\maketitle
\begin{abstract}
  The remarkable advancement in text-to-image generation models significantly boosts the research in ID customization generation. However, existing personalization methods cannot simultaneously satisfy high fidelity and high-efficiency requirements. Their main bottleneck lies in the prompt image encoder, which produces weak alignment signals with the text-to-image model and significantly increased model size. Towards this end, we propose a lightweight Inv-Adapter, which first extracts diffusion-domain representations of ID images utilizing a pre-trained text-to-image model via DDIM image inversion, without additional image encoder. Benefiting from the high alignment of the extracted ID prompt features and the intermediate features of the text-to-image model, we then embed them efficiently into the base text-to-image model by carefully designing a lightweight attention adapter. We conduct extensive experiments to assess ID fidelity, generation loyalty, speed, and training parameters, all of which show that the proposed Inv-Adapter is highly competitive in ID customization generation and model scale.
\end{abstract}

\section{Introduction}
\label{sec:intro}
%
Recent advanced models like DALL-E2~\cite{ramesh2021zero}, Imagen~\cite{saharia2022photorealistic}, Stable Diffusion(SD)~\cite{rombach2022high}, and SDXL~\cite{podell2023sdxl} have demonstrated remarkable progress~\cite{radford2021learning,raffel2020exploring,peebles2023scalable,schuhmann2022laion,schuhmann2021laion}, facilitating the emergence of tasks centered on identity (ID) customization generation.
%
%
This field is vital for personalized photo creation, animation, and e-commerce and thus attracts  comprehensive research attention.

ID customization generation requires finer details, subtler semantic distinctions, and higher fidelity~\cite{liang2024caphuman,wang2024instantid}, setting it apart from the standard controllable generation tasks~\cite{zhang2023adding,mou2023t2i}.
A popular direction is fine-tuning text-to-image models, such as Text-inversion~\cite{gal2022image}, DreamBooth~\cite{ruiz2023dreambooth}, and Low-Rank Adaptation (LoRA)~\cite{chen2023photoverse}.
%
%
Despite their prevalent applications within the research community, these methods exhibit notable limitations (requirements of multiple ID-consistent images, significant computational resources) in real-world scenarios. 
%

\begin{figure}[t]
    \centering\setlength{\abovecaptionskip}{0.1cm}
    \includegraphics[width=.99\linewidth]{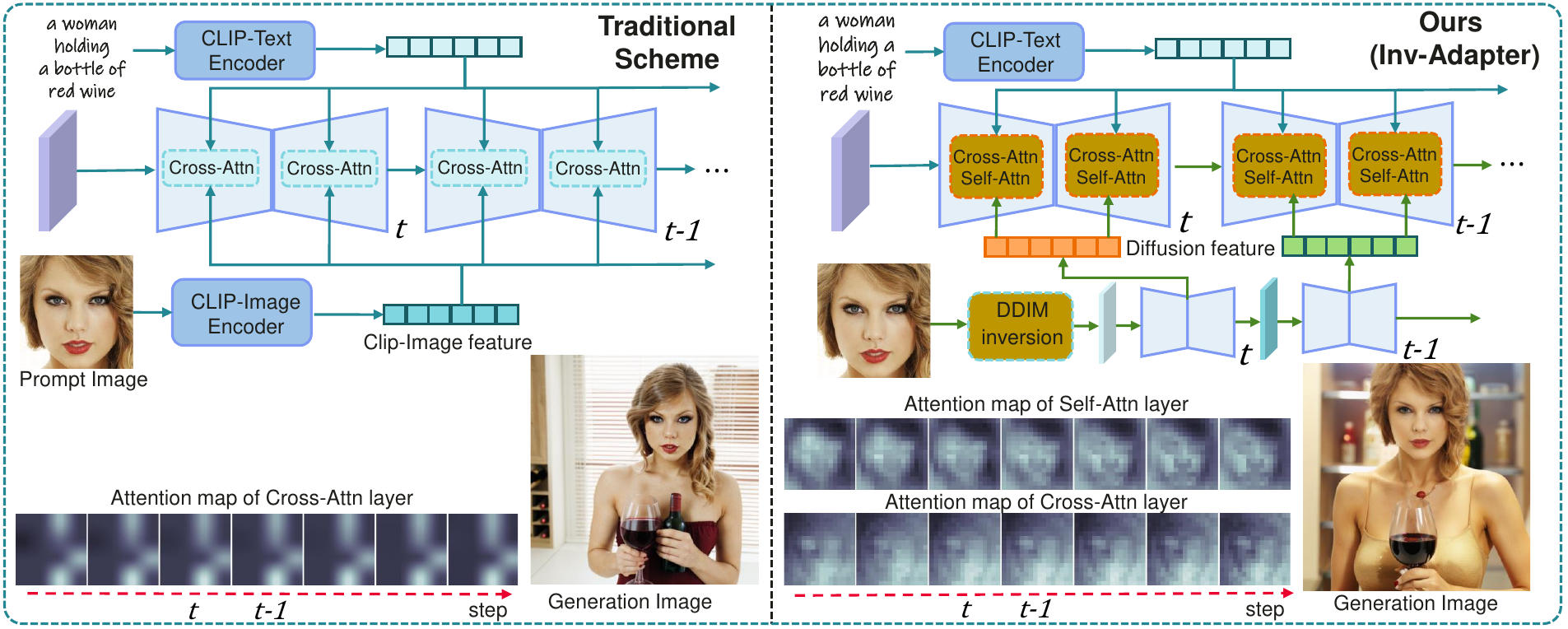}
    
    \caption{Comparison of popular scheme (IP-Adapter~\cite{ye2023ip}) with the proposed Inv-Adapter. 
    %
    From attention maps, the proposed Inv-Adapter captures detailed face features more accurately.
    }
    \label{fig1}
    \vspace{-0.4cm}
\end{figure}

To improve the identity learning ability from prompt images while retaining their inherent generation capability, efficient training methods~\cite{wei2023elite,ye2023ip,mou2023t2i} have received much attention. 
Unlike fine-tuning methods, they are free of fine-tuning in the inference stage, and rely only on one image with a single denoising process for ID customization generation. 
As in Photomaker~\cite{li2023photomaker}, ID preservation is achieved by optimizing the image feature extractor and utilizing the pre-trained cross attention layer to fuse the text, image, and intermediate features.
%
%
IP-Adapter~\cite{ye2023ip} is a recent popular ID preservation scheme, which extracts prompt image features by CLIP~\cite{radford2021learning} image encoder, and then integrates them via an additional cross attention layer.

 After analyzing above methods~\cite{ye2023ip,li2023photomaker}, it is evident that two factors play a vital role in ID customization generation: 1) how to extract highly aligned and detailed face representations with the diffusion model, and 2) how to efficiently integrate these features into the diffusion model.
However, previous methods fail to adequately explore the potential hidden in these aspects.
\begin{wraptable}{r}{0.6\textwidth}
\centering
\vspace{-0.03cm}
\scriptsize
\setlength{\abovecaptionskip}{0cm}
\setlength{\belowcaptionskip}{-0.01cm}
\caption{Comparison of the proposed lightweight inv-Adapter with recent approaches.}
\begin{tabular*}{0.99\linewidth} {@{\extracolsep{\fill}}c|c}
\hline
Method                 & Additional Image Encoder \\ \hline
IP-Adapter-Plus~\cite{ye2023ip}        & CLIP-ViT-H-14~\cite{ilharco_gabriel_2021_5143773}                                      \\
IP-Adapter-FaceID-Plus~\cite{ye2023ip}  & face model (buffalo\_l)~\cite{Deng2020CVPR}, CLIP-ViT-H-14~\cite{ilharco_gabriel_2021_5143773}             \\
Photomaker~\cite{li2023photomaker}             & CLIP-ViT-L-14~\cite{radford2021learning}                                       \\
InstantID~\cite{wang2024instantid}              & face model (antelopev2) ~\cite{Deng2020CVPR}, IdentityNet~\cite{wang2024instantid}    \\
Ours                   & \textbf{None }                                                     \\ \hline
\end{tabular*}\label{tab:intro}
\vspace{-0.3cm}
\end{wraptable}
First, they typically rely on the additional image encoder~\cite{radford2021learning}, and the extracted features are weakly aligned with the text-to-image model.
The most likely reason is that extracted image features (\textit{e.g.}, CLIP image features) have a modality gap with intermediate features of the diffusion model and cannot be inserted into the text-to-image models with high fidelity.
As shown in Figure~\ref{fig1}, we visualize the attention maps in different steps of the face prompt image. It clearly indicates that only part of the face information can be attended to the IP-Adapter~\cite{ye2023ip}.
Second, these methods usually merge image features in the cross attention layer only. They ignore the critical role of the self attention mechanism.
Beside, the additional image encoder significantly increases the model scale, which is not conducive to end-side deployment, as shown in Table~\ref{tab:intro}.
Toward this end, we introduce Inv-Adapter, a high-fidelity yet lightweight approach to ID customisation through image inversion and lightweight adapter.
The main contributions of this work are three-fold:
First, we \textit{discard} the additional CLIP image encoder and instead \textit{reuse} the pre-trained text-to-image model.
Specifically, we introduce DDIM image inversion~\cite{song2020denoising} and extract prompt image representations in the diffusion domain (called diffusion features) at each step in the denoising process.
Since the face diffusion features and the intermediate features of the text-to-image process come from the same model, our diffusion features can be more effectively embedded in the text-to-image model.
%
Second, benefiting from the consistency of the diffusion features with the intermediate features, we propose a lightweight Embedded Attention Adapter (only $48$M parameters) that efficiently injects the diffusion features into both the self and cross attention layers simultaneously.
Figure~\ref{fig1} shows that the self attention and cross attention layers can exploit more detailed information about ID, significantly boosting generative fidelity. 
%
Finally, unlike previous works using fixed image embeddings, our framework accompanies a denoising process to extract and inject diffusion features at each step. As in the self attention layer of Figure~\ref{fig1}, with the gradual denoising process, more detailed face information is explored.
Extensive experiments demonstrate that our proposed lightweight Inv-Adapter achieves advanced performance in generating faithful, detailed, and high-fidelity images.
%


%

%
\begin{figure}[h]
    \centering\setlength{\abovecaptionskip}{0.05cm}
    \includegraphics[width=0.99\linewidth]{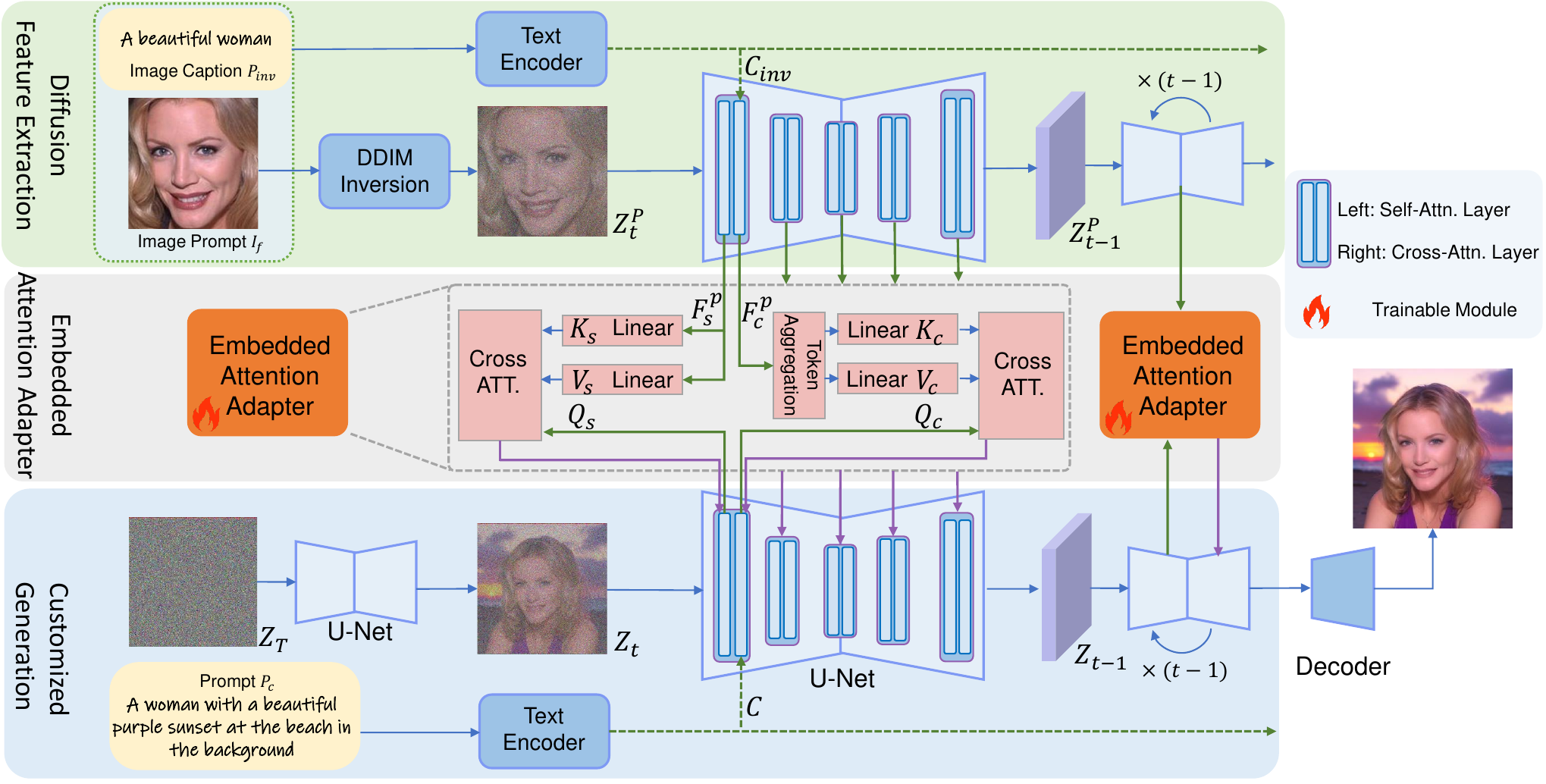}
    \caption{Overview of the proposed Inv-Adapter. 
    First, in the diffusion feature extraction phase, the latent noise $Z_t^p$ is obtained by DDIM inversion of prompt image $I_f$. 
    %
    %
    Then, in the denoising process on latent noise $Z_t^p$, we extract diffusion features, which are the intermediate representations of the diffusion model and contain detailed information about $Z_t^p$. 
    The diffusion features are inserted into the same pre-trained text-to-image through the Embedded Attention Adapter (EAA) to generate the image that preserves ID information. 
    %
    %
    Finally, the final result that aligns the prompts $P_c$ and $I_f$ is obtained after iterating the $T$ steps. }
    \label{fig2}
    \vspace{-0.6cm}
\end{figure}

\section{Related Work}
\label{sec:related}

%

\myparagraph{Image Inversion.}
Image inversion aims to invert an image into the latent noise with the expectation that this latent noise can generate the original image, which is mainly used for real image editing~\cite{mokady2023null}. 
The inversion technique has been widely studied in GANs-based image editing~\cite{creswell2018inverting,bermano2022state,zhu2016generative,dalva2022vecgan}.
In the era of diffusion models, similar diffusion model-based image inversion approaches have also been developed.
%
%
Image inversion with DDIM sampling~\cite{song2020denoising} based on a deterministic ODE process has received wild attention.
%
%
%
In this paper, for ID customization generation, we leverage DDIM inversion to obtain the latent noise of the corresponding prompt image and use the latent noise to extract diffusion features via a pre-trained text-to-image diffusion model.

\myparagraph{Personalization.}
As text-to-image models exhibit excellent diverse generation capabilities, recent works have begun to explore controllable generation.
These researches primarily bifurcate into two streams: contour condition-based generation (\textit{e.g.}, pose) like ControlNet~\cite{zhang2023adding} and T2I-Adapter~\cite{mou2023t2i}, and ID customization generation aiming to preserve the face ID information and simultaneously allow diverse generation~\cite{ruiz2023dreambooth,li2023photomaker,ye2023ip,wei2023elite}. 
ID customization generation was initially pioneered by methodologies such as Dreambooth~\cite{ruiz2023dreambooth} and Textual Inversion~\cite{gal2022image}. 
These approaches necessitate model fine-tuning for each identity during the inference phase, which is typically inconvenient and resource-intensive~\cite{ruiz2023dreambooth}.
To circumvent these limitations, some works introduce adapter~\cite{mou2023t2i,ye2023ip}, which integrates additional trainable modules while maintaining the generation ability of the base pre-trained text-to-image models~\cite{houlsby2019parameter,mou2023t2i}. 
%
 %

Innovative approaches such as IP-Adapter~\cite{ye2023ip}, PhotoVerse~\cite{chen2023photoverse}, and PhotoMaker~\cite{li2023photomaker} have been introduced for ID customization generation task. 
 IP-Adapter~\cite{ye2023ip} employs a CLIP image encoder for facial feature extraction and relies on the cross attention layers to embed facial identity information into generated images. 
  PhotoVerse~\cite{chen2023photoverse} maps image prompt features to the image and text domains and then incorporates additional cross attention layers to insert them into diffusion models.
 PhotoMaker~\cite{li2023photomaker} leverages stacked embeddings of multiple facial images processed through CLIP image encoder, and integrates these embeddings into the diffusion model via pre-trained cross attention layers to ensure identity preservation.
However, the above approaches ignore the gap between the CLIP visual representation and the intermediate feature domain of the diffusion model.
Concurrent work InstantID~\cite{wang2024instantid} reveals that a plain face feature extractor outperforms the CLIP image encoder, but the domain gap compared to the intermediate feature domain still exists.
Differently, the proposed method extracts ID diffusion embedding without additional image encoder, which is straightforward yet effective in ID customization generation.


\section{Method}
\label{sec:method}

%
%
The overall framework of Inv-Adapter is shown in Figure~\ref{fig2}, which contains diffusion feature extraction, embedded attention adapter (EAA), and customized generation. 
%
%
%
%
%

\subsection{Diffusion Feature Extractor}
Previous works typically utilize the CLIP image encoder~\cite{radford2021learning} or pre-trained face model~\cite{ilharco_gabriel_2021_5143773,wang2024instantid} to extract image representations of the prompt image.
%
However, they require additional image encoders, significantly increasing end-side deployment costs.
In addition, the base text-to-image model fails to utilize these image features effectively due to the modality gap. 
\begin{figure*}[t]
    \centering\setlength{\abovecaptionskip}{1pt}
    \includegraphics[width=0.99\linewidth]{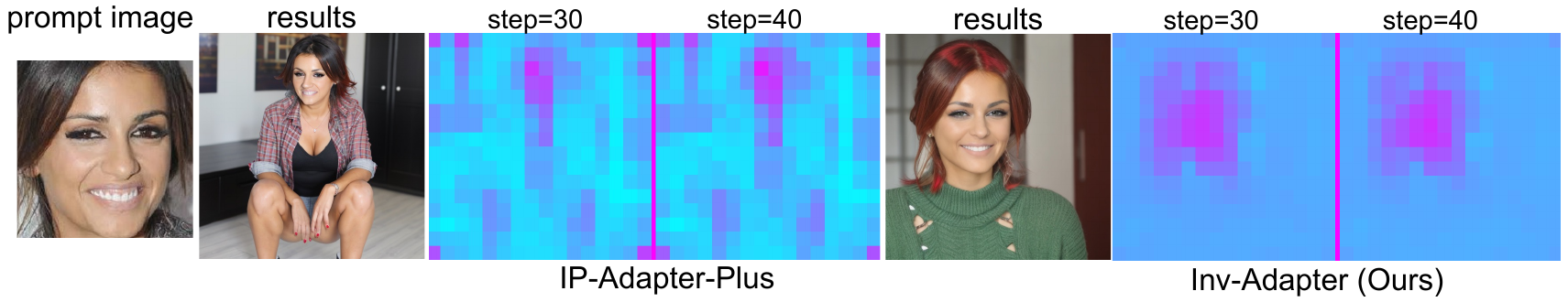}
    \caption{%
    %
    %
    Visualization of the attention maps of the generated results with the prompt images. 
    %
    %
    Our proposed Inv-Adapter employing diffusion features makes the prompt images focus only on the critical face region, which is ideal. 
    }
    \label{fig3}
    \vspace{-0.4cm}
\end{figure*}
%
%
As illustrated in Figure~\ref{fig3}, we visualize attention maps of the prompt images, which represent the attention of the prompt image to the generated image.
Image features extracted by IP-Adapter-Plus~\cite{ye2023ip} may be suboptimal, which leads to the prompt image failing to focus only on the face region.

In this paper, we propose an ID feature extraction method based on a pre-trained text-to-image model without additional custom CLIP image encoder.
%
The prompt image $I_f$ is first encoded as an initial latent noise $P_0$, and $T$ steps of the denoising process are performed based on the DDIM inversion~\cite{dhariwal2021diffusion} in the opposite direction using a pre-trained text-to-image diffusion model, \textit{i.e.}, $P_0 \rightarrow P_T$ instead of $P_T \rightarrow P_0$. 
$P_T$ is denoted as the latent noise corresponding to $I_f$. 
%
%
In Inv-Adapter, we set the classifier-free guidance parameter as $1$ to get high quality inversion results (More descriptions in the appendix~\ref{supp}). 
The latent noise $P_T$ of the prompt image performs a normal denoising process, where the corresponding image caption $P_{inv}$ serves as the text prompt. 
During this denoising process, we extract intermediate features $F_s^p$ in the self attention layer and $F_c^p$ in the cross attention layer. 
Then, we regard $F_s^p$ and $F_c^p$ as ID diffusion features and inject them into the customized generation branch. 
%
%
%

\subsection{Embedded Attention Adapter}
\begin{figure}[t]
    \centering
    \includegraphics[width=.99\linewidth]{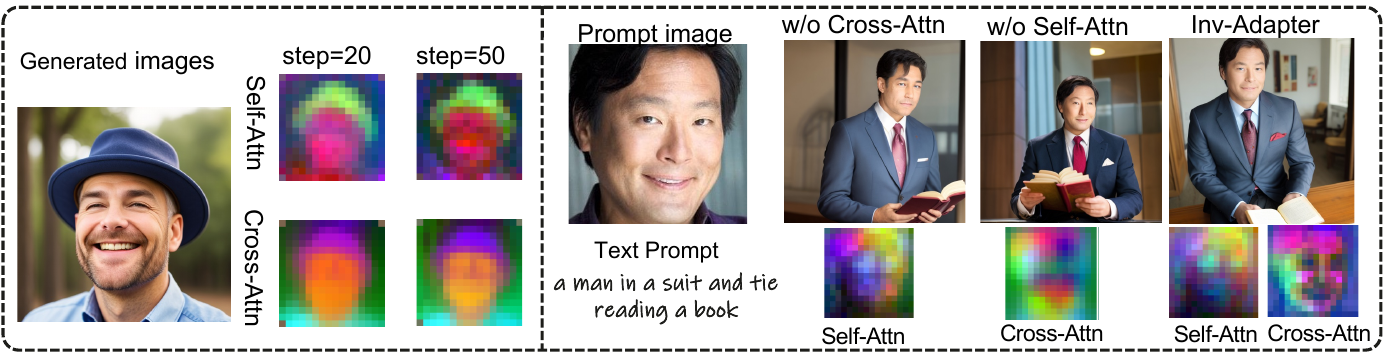}
    \vspace{-0.25cm}
    \caption{Left: the image generated by the SD model~\cite{rombach2022high} and the self attention and the cross attention maps in different steps. 
    Right: generated results and attention visualisation of the proposed Inv-Adapter ablation experiment on the attention layers.
    %
    }
    \label{fig4}
    \vspace{-0.4cm}
\end{figure}

%
The denoising network is a U-Net consisting of multiple blocks, each containing a self attention layer and a cross attention layer~\cite{rombach2022high}.
As shown on the left of Figure~\ref{fig4}, we visualize the attention maps from the self attention layers and cross attention layers at different steps.
These results demonstrate that both the self attention and cross attention layers are critical to the generated images' structure, style, and content.
Nevertheless, previous ID customization generation methods~\cite{ye2023ip,wang2024instantid} mostly ignore the importance of self attention layer. 
Thanks to the consistency of the diffusion features with the intermediate features in the text-to-image generation process, the proposed Inv-Adapter inserts face diffusion features to both self and cross attention layers simultaneously.

We describe the process of the Embedded Attention Adapter by taking the diffusion process at the $t$-th step (\textit{i.e.}, $Z_t \rightarrow Z_{t-1}$) as an example. 
In the original SD~\cite{rombach2022high}, given the input features $F_s$ of self attention layer, $F_c$ of cross attention layer, and the text features $C$, the output $O_s$/$O_c$ of the self/cross attention layer are represented as follows:
\begin{equation}
    O_s = Softmax(\frac{F_sW_q^s\cdot(F_sW_k^s)^T}{\sqrt{d}})F_sW_v^s, O_c = Softmax(\frac{F_cW_q^c\cdot(CW_k^c)^T}{\sqrt{d}})CW_v^s,
\end{equation}
where $W_q^s,W_k^s,W_v^s$/$W_q^c,W_k^c,W_v^c$ denote the weight matrices of the trainable linear projection layers in self/cross attention layer, respectively. $Softmax(\cdot)$ denotes the softmax function in the last dimension.

%
%
%
%

In the self attention layer, first, diffusion feature $F_s^p$ of the prompt image is projected into the key and value matrices of the attention operation through two independent linear projection layers, denoted as $K_s$ and $V_s$, respectively. 
%
%
Then, $F_s$ is projected into a query matrix $Q_s$ by the pre-trained weight matrix $W_q^s$.
Next, the output $O_s^p$ of the Embedded Attention Adapter is obtained by the cross attention operation and is denoted as follows:
\begin{equation}
    O_s^p = Attention(Q_s,K_s,V_s) =Softmax(\frac{Q_s(K_s)^T}{\sqrt{d}})V_s.
\end{equation}
Finally, we perform the feature fusion before the MLP layer~\cite{dosovitskiy2020image}. 
Specifically, we merge the output $O_s$ of the original attention layer with the output $O_s^p$ of the EAA as follows:
\begin{equation}
    O_s^{\prime}= O_s +\sigma \cdot O_s^p,
\end{equation}
where $ O_s^{\prime}$ represents the final output and $\sigma$ denotes the fusion hyperparameter, which represents the influence degree of $O_s^{\prime}$ on the customized generation branch.

In the cross attention layer, the imbalance number of prompt image tokens and text tokens is a non-negligible issue.
We empirically find that using original-scaled diffusion features may reduce the impact of textual prompts, diminishing the editability and fidelity of the generation. 
To this end, we employ a token aggregation operation for diffusion features at the beginning of the cross attention layer.
Specifically, for the extracted diffusion feature $F_c^p \in \mathcal{R}^{b\times s \times d}$, it is directly mapped to aggregation feature $F_c^{p^{\prime}} \in \mathcal{R}^{b\times s_c \times d}$ through a linear layer, where $s$ and $s_c$ represent the sequence length of the original features and the aggregation features, respectively.
%
Then, the aggregation feature $F_c^{p^{\prime}}$ is mapped to the key matrix $K_c$ and value matrix $K_v$ of the attention operation through two separate linear layers.
$F_c$ is projected to the query matrix $Q_c$ of the attention operation through the original linear layer.
Then, we obtain the output of the Embedded Attention Adapter in the cross attention layer as follows:
\begin{equation}
    O_c^p = Attention(Q_c,K_c,V_c) =Softmax(\frac{Q_c(K_c)^T}{\sqrt{d}})V_c.
\end{equation}
Finally, similar to EAA at the self attention layer, the embedding fusion operation is performed before the MLP layer.
\begin{equation}
    O_c^{\prime}= O_c +\sigma \cdot O_c^p,
\end{equation}
where $ O_c^{\prime}$ represents the fusion output in the cross attention layer.

\subsection{Model Training} \label{sec3.3}
We train Inv-Adapter with image-text pairs with face prompt images. 
Specifically, we freeze the parameters of the pre-trained text-to-image generation model and merely optimize the parameters of the linear layers of the Embedded Attention Adapter.
First, for each image-text pair~$(I, P)$, where $I$ denotes the prompt image and $P$ denotes image caption~$P_{inv}$, we crop the image $I$ with a square box based on the face detection algorithm~\cite{guo2021sample,Deng_2020_CVPR} to obtain face prompt image $I_f$. 
During the training phase, the conditional text prompt $P_c$ is the same as the image caption $P_{inv}$. In the inference phase, $P_c$ represents the text prompt.
Then, by DDIM inversion, the latent noise $Z_t^p$ of face image $I_f$ at step $t$ can be efficiently obtained.
Finally, the Inv-Adapter is trained using the following optimization objective.
\begin{equation}
    \mathcal{L}=\mathbb{E}_{x_0,\epsilon\sim \mathcal{N}(0,1)  }\left \|\epsilon-\epsilon_\theta {(Z_t,C,Z_t^p,C_{inv},t)}  \right \|_2, 
\end{equation}
where $C$ and $C_{inv}$ represent text features extracted by $P_c$ and $P_{inv}$ through the CLIP text encoder, $t$ denotes the denoising step, $\theta$ represents the parameters of EAA, $\epsilon_\theta(\cdot)$ represents the overall model. 

%

\section{Experiments}
\label{sec:Experiments}
\subsection{Experimental Setup}\label{sec4.1}

\begin{table*}[h!]
\setlength{\abovecaptionskip}{0cm}
\setlength{\belowcaptionskip}{-0.01cm}
\scriptsize\renewcommand{\arraystretch}{1}
\caption{Comparison of quantitative experimental results of the proposed Inv-Adapter with recent state-of-the-art methods on the Sample-1K dataset. The proposed method achieves state-of-the-art performance for face fidelity and image quality.}
\begin{tabularx}{0.99\textwidth}{c|c|cc|ccc|cc}
\hline
\multirow{1}{*}{Method}  &Prompt& {BLIP}($\uparrow$)  & {CLIP-T} ($\uparrow$) & {CLIP-I} ($\uparrow$) & {DINO} ($\uparrow$) & {SIM} ($\uparrow$) & {IQA} ($\uparrow$) & {FID} ($\downarrow$)    \\ \hline
IP-Adapter~\cite{ye2023ip}   & Full                                                                    & 74.18                 & 27.09                   & 69.23                   & 42.11                 & 30.07                     & 84.26                    & 253.48               \\
IP-Adapter-Plus~\cite{ye2023ip}   & Face                                                                    & 74.22                 & 27.01                   & 67.23                   & 42.97                 & 37.45                     & 86.65           & 285.20                \\
IP-Adapter-FaceID-Plus~\cite{ye2023ip}&Face&78.85&27.43&65.52&58.12&63.11&86.15&230.51\\
PhotoMaker~\cite{li2023photomaker}        & Full                                                                   & {80.46} &    28.28                    &   67.44                        &           41.76              &   26.85                                             &                  68.57        &265.23  \\ 
InstantID~\cite{wang2024instantid}& Face & 81.12 &28.33 & 70.25 & 62.53 & 67.22&84.11& 231.45 \\ \hline
Inv-Adapter(Ours) & Face                                                                    & {}79.65        & {28.45 }         & {71.81}          & {64.33}        & {65.12}            & {86.80 }                   & {228.10}   \\    \hline
\end{tabularx}\label{tab1}
\vspace{-0.2cm}
\end{table*}

\begin{table}[h]
\scriptsize\renewcommand{\arraystretch}{1}
\setlength{\abovecaptionskip}{0cm}
\setlength{\belowcaptionskip}{-0.01cm}
\caption{Comparison of quantitative experimental results of the proposed Inv-Adapter with recent state-of-the-art methods on the Celebrity dataset.}
\begin{tabularx}{0.99\textwidth}{c|c|cc|ccc|ccc}
\hline
\multirow{1}{*}{Method}  &Prompt& {BLIP}($\uparrow$)  & {CLIP-T} ($\uparrow$) & {CLIP-I} ($\uparrow$) & {DINO} ($\uparrow$) & {SIM} ($\uparrow$) & {IQA} ($\uparrow$) & {FID} ($\downarrow$)    \\ \hline
IP-Adapter~\cite{ye2023ip}        & Full                                                                    & 71.95                      & 24.86                                       & 71.21                                       & 52.01                                     & 28.91                                        & 84.92                                        & 252.56              \\
IP-Adapter-Plus~\cite{ye2023ip}   & Face                                                                    &  76.37                     & 27.07                                       & 67.68                                       & 50.45                                     & 40.45                                         & 87.41                               & 220.58               \\
IP-Adapter-FaceID-Plus~\cite{ye2023ip}&Face&79.15&27.14&74.69&57.19&{68.08}&89.26&209.86\\
PhotoMaker~\cite{li2023photomaker}        & Full                                                                    & {80.40}                      & {29.26}                                       & 70.08                                       & 41.52                                     & 34.62                                         & 70.30                                        & 244.25              \\ 
InstantID~\cite{wang2024instantid}& Face & 81.31 &29.68 & 72.69 & 57.67 & 69.83&86.42& 215.03 \\ \hline
Inv-Adapter(Ours) & Face                                                                    & 80.04             & {28.21}               & {{76.79}}               & {{61.49}}             & 65.42              &{{89.56} }                        & {201.38}       \\      \hline
\end{tabularx}\label{tab2}
\vspace{-0.2cm}
\end{table}

\begin{table}[h!]
\scriptsize
\setlength{\abovecaptionskip}{0cm}
\setlength{\belowcaptionskip}{-0.01cm}
\scriptsize\renewcommand{\arraystretch}{1.2}
\caption{Comparison of trainable parameters, training datasets, model scale, and generation speed of the proposed method with recent state-of-the-art methods.}
\resizebox{0.99\linewidth}{!}{
\begin{tabular}{c|ccccc}
\hline
  & {IP-Adapter-Plus\cite{ye2023ip}} & IP-Adapter-FaceID-Plus\cite{ye2023ip}&{PhotoMaker\cite{li2023photomaker}} & {InstantID\cite{wang2024instantid}}        & {Inv-Adapter} \\
                                                                     \hline
Train. Param($\downarrow$)                           & 49M   &78M                           & 460M                        & $\gg$1024M & \textbf{48M}                                \\
\hline
Training Data (K)&300&300&112&50000&170 \\ \hline
Model Scale($\downarrow$)&1.13G & 2.49G & 8.39G &$\gg$10G &\textbf{1.13G} \\ \hline
Speed($\downarrow$)                                          & \textbf{4.5s} &5s                              & 14s                         &$\gg$14s   & 11s                                \\ \hline
\end{tabular}
}\label{tab3}
\vspace{-0.2cm}
\end{table}

\myparagraph{Training Dataset.}
We train Inv-Adapter on two publicly available datasets including CelebA-HQ~\cite{karras2018progressive} and FFHQ~\cite{karras2019style}. The overall training dataset contains 170k images.

\myparagraph{Evaluation Dataset.}
(1) Sample-1K dataset: due to the lack of public evaluation benchmarks, we sample 1k images from  CelebA-HQ~\cite{karras2018progressive} and FFHQ~\cite{karras2019style} datasets as an evaluation dataset, denoted as Sample-1K in the following experiments. 
(2) Celebrity dataset: images of celebrities manually collected from the Internet.
To evaluate the zero-shot ID customization generation capability, the training datasets and evaluation datasets do not contain the same images. 
For each prompt image, we construct $42$ text prompts and generate $5$ images for each text prompt.

\myparagraph{Compared Baseline Methods.} (1)~IP-Adapter~\cite{ye2023ip}: its backbone is SD $v1.5$ and the prompt image is the original image. 
(2)~IP-Adapter-Plus~\cite{ye2023ip}: its backbone is SD $v1.5$ and the prompt image is a face image. 
%
(3)~IP-Adapter-Faceid-Plus~\cite{ye2023ip}: its backbone is SD $v1.5$ and it utilizes a pre-trained face feature extractor from the Insightface library~\cite{deng2018arcface} and CLIP image necoder to extract the ID features of the prompt image as embedded features (v2 version).
(4)~PhotoMaker~\cite{li2023photomaker}: its backbone is SDXL-1.0 and trained with a self-collected dataset including a lot of celebrities.
(5)~InstantID~\cite{wang2024instantid}: its backbone is SDXL-1.0, combining IP-Adapter and a huge IdentityNet ( ControlNet~\cite{zhang2023adding}). 

\myparagraph{Metrics.}
To comprehensively evaluate the proposed Inv-Adapter, we consider three aspects of metrics.
(1)~\textit{Loyalty.} 
We employ CLIP-T~\cite{radford2021learning} and BLIP~\cite{li2023blip,chefer2023attend} scores to evaluate the similarity of text prompts and the generated images.
(2)~\textit{ID~fidelity.}
 We first report the DINO~\cite{caron2021emerging} and CLIP-I~\cite{gal2022image} scores of the face image $I_f$ with the generated image, respectively. 
Second, to evaluate face fidelity more accurately, we extract the face features of the generated and prompt images by Insightface~\cite{deng2018arcface} and compute their similarity, denoted as FACE-SIM.
(3)~\textit{Quality.}
 We introduce CLIP-IQA~\cite{wang2023exploring} and FID~\cite{heusel2017gans}. The former utilizes the strong prior information of CLIP to evaluate the look and feel.

\myparagraph{Implementation Details.}
During the training phase, we use Stable Diffusion (SD $v1.5$) as the backbone of Inv-Adapter. 
We fix all parameters of SD $v1.5$ and only train the linear projection layers of the embedded attention adapter. We crop face images using the insightface library~\cite{deng2018arcface}. 
For image inversion, DDIM inversion is used with classifier-free guidance, and the guidance scale is set to 1. 
The proposed Inv-Adapter is optimized with AdamW~\cite{loshchilov2017decoupled} on 16 NVIDIA V100 GPUs for one week with a batch size of $4$. We set the learning rate as $1e-4$, weight decay as $1e-2$, and $\sigma$ as $1$. 
During the inference phase, we can employ community models to achieve more realistic image generation. In this paper, we use the \textit{Realistic Vision V4.0} model from \textit{huggingface}. We adopt the DDIM sampler with 50 steps. The classifier-free guidance scale is set to 7.5, and $\sigma$ can be set to $[0.6, 1]$.

\begin{figure}[h]
    \centering\setlength{\abovecaptionskip}{0.1cm}
    \includegraphics[width=0.99\linewidth]{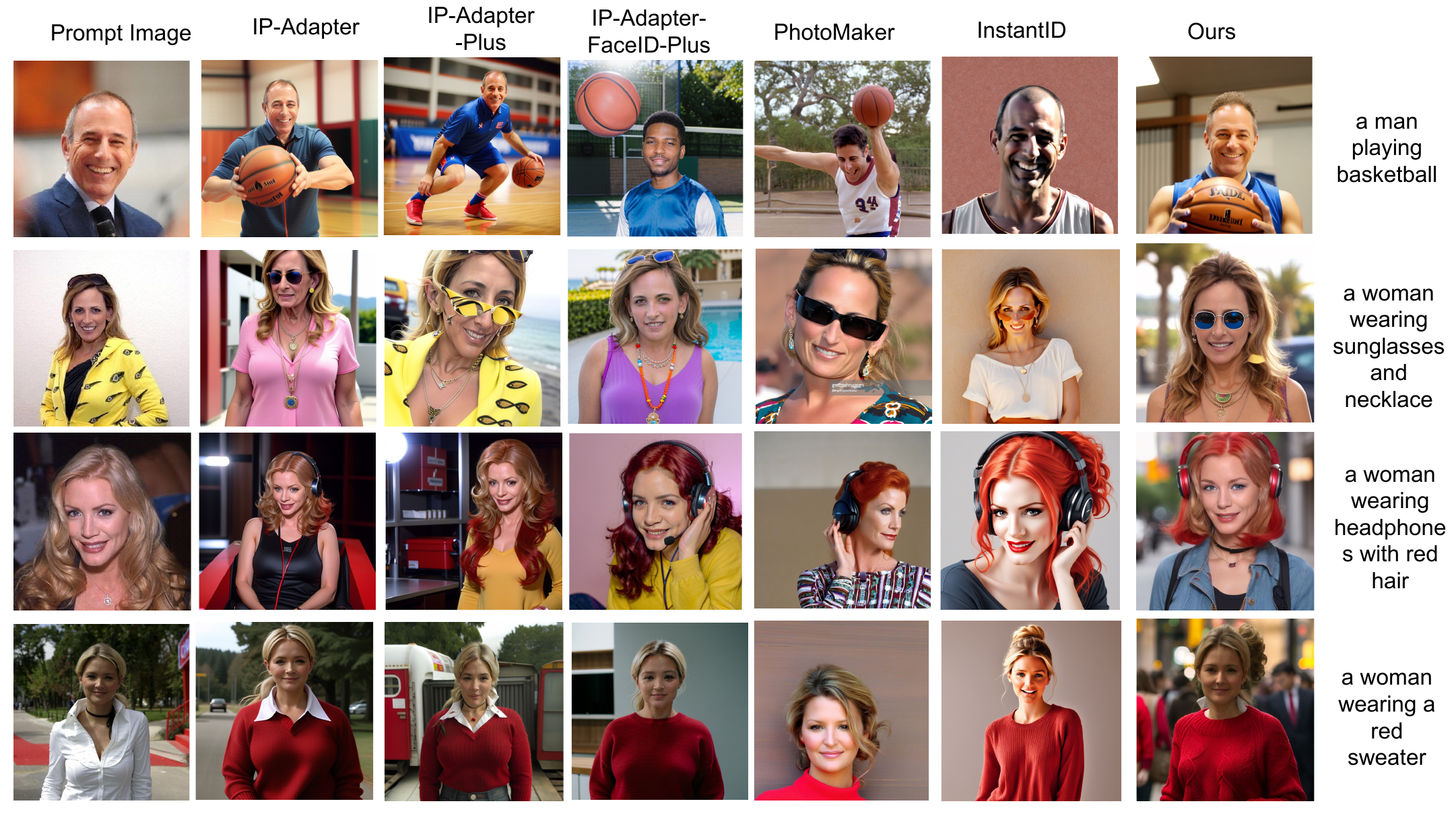}
    \caption{Comparison results of Inv-Adapter with recent advanced IP-Adapter~\cite{ye2023ip}, IP-Adapter-Plus~\cite{ye2023ip}, IP-Adapter-FaceID-Plus~\cite{ye2023ip}, PhotoMaker~\cite{li2023photomaker}, and InstantID~\cite{wang2024instantid} on Sample-1K.}
    \label{fig5}
    \vspace{-0.5cm}
\end{figure}
\subsection{Results}

\myparagraph{Quantitative Result.}
We report the quantitative results of Inv-Adapter compared with recent state-of-the-art methods on the Sample-1K and the Celebrity datasets in Tables \ref{tab1} and Table \ref{tab2}, respectively. 
First, in terms of loyalty, our method exhibits significantly higher scores than IP-Adapter~\cite{ye2023ip} and close to PhotoMaker~\cite{li2023photomaker}. 
This shows that the proposed method provides a better trade-off between text and image prompts
Second, regarding the three metrics of ID fidelity, CLIP-I, DINO, and FACE-SIM, the proposed method significantly outperforms the previous methods.
These experimental results demonstrate that the face diffusion features extracted by Inv-Adapter possess a higher degree of alignment with the pre-trained text-to-image model, which facilitates rich face details preservation. 
In addition, Inv-Adapter achieves higher FACE-SIM scores in both Table~\ref{tab1} and Table~\ref{tab2}, marking that our method can generalize well to unlearned ID for customized generation.
Finally, the CLIP-IQA and FID scores show that the Inv-Adapter can generate high-quality images.

\myparagraph{Qualitative Result.}
Figure~\ref{fig5} shows the generation results of the proposed Inv-Adapter and recent state-of-the-art methods for unlearned face scenarios. 
The visualization results indicate that IP-Adapter~\cite{ye2023ip} loses more face details but loyally obeys the textual prompts. 
IP-Adapter-Plus~\cite{ye2023ip} captures more face details, but there is still a large gap with our Inv-Adapter. 
PhotoMaker~\cite{li2023photomaker} performs poorly for the unlearned face prompt image as the ID embedding fails to summarize the rich detail information from a single ID image. 
InstantID~\cite{wang2024instantid} outperforms the above methods, but still has a gap in the face details compared to the proposed method. 
Inv-Adapter performs optimally because the extracted face diffusion features contain rich, detailed information and can be effectively embedded in the text-to-image model. 
More importantly,  the base model of our proposed method is \textit{SD~$v1.5$}, which already outperforms InstantID~\cite{wang2024instantid} and Phtotomaker~\cite{li2023photomaker} based on  SDXL-1.0.

\begin{figure}[t]
    \centering\setlength{\abovecaptionskip}{0.1cm}
    \includegraphics[width=0.99\linewidth]{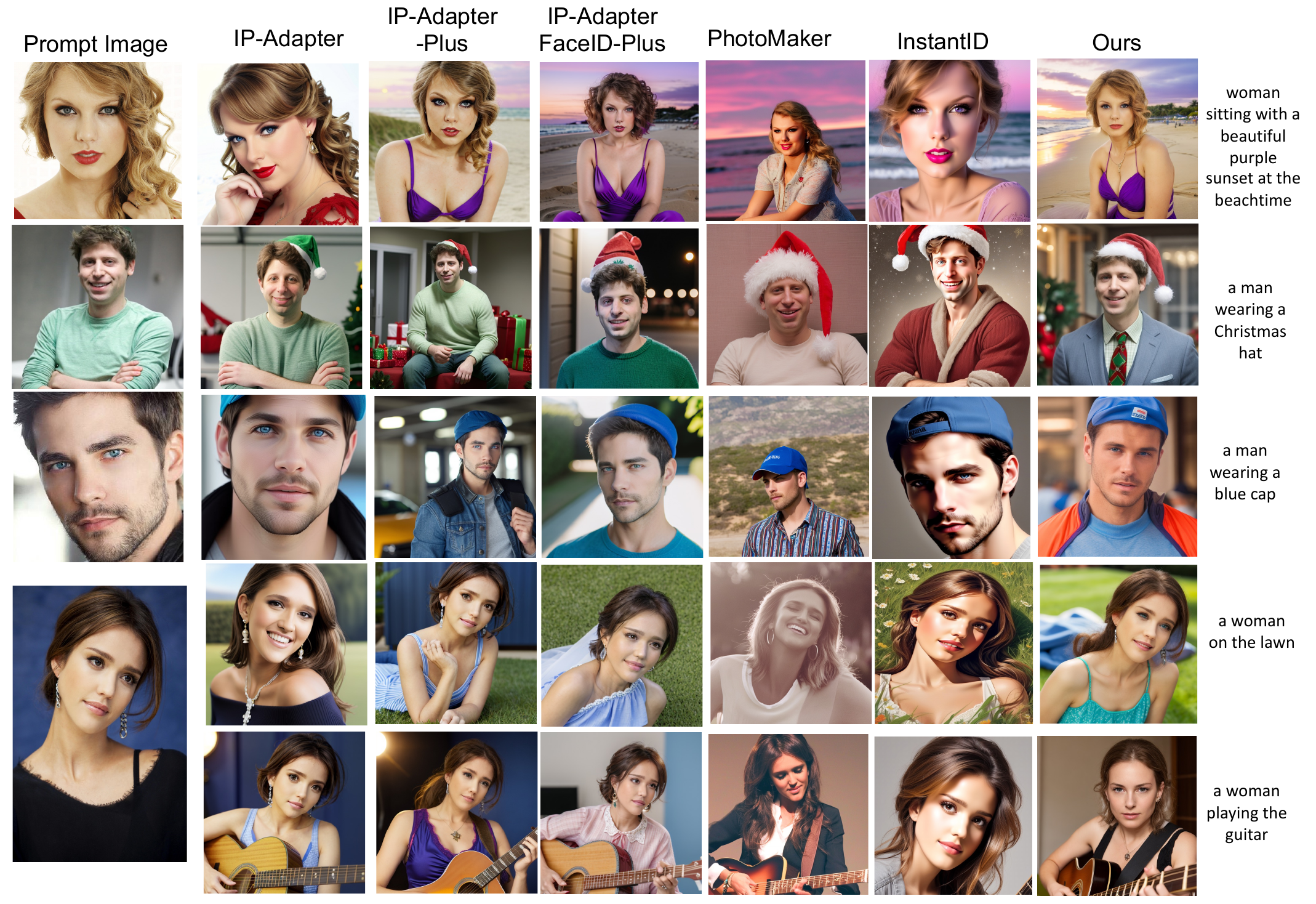}
    \caption{Comparison results of Inv-Adapter with recent advanced IP-Adapter~\cite{ye2023ip}, IP-Adapter-Plus~\cite{ye2023ip}, IP-Adapter-FaceID-Plus~\cite{ye2023ip},  PhotoMaker~\cite{li2023photomaker}, and InstantID~\cite{wang2024instantid} on Celebrity dataset.}
    \label{fig6}
    \vspace{-0.0cm}
\end{figure}

As shown in Figure~\ref{fig6},  we demonstrate the generation results of the proposed method compared to the state-of-the-art methods on the Celebrity dataset.
PhotoMaker~\cite{li2023photomaker} shows better ID preservation for celebrity images than for ordinary faces from Sample-1K.
A possible reason is that the Photomaker~\cite{li2023photomaker} involves celebrity images (\textit{e.g.}, Taylor Swift) for model training. 
%
%
Compared with SDXL-based InstantID~\cite{wang2024instantid}, our Inv-Adapter based on SD $v1.5$ is competitive in terms of ID fidelity and loyalty in the Celebrity dataset.
%
%
%

\begin{figure*}[t!]
    \centering\setlength{\abovecaptionskip}{0.1cm}
    \includegraphics[width=0.99\linewidth]{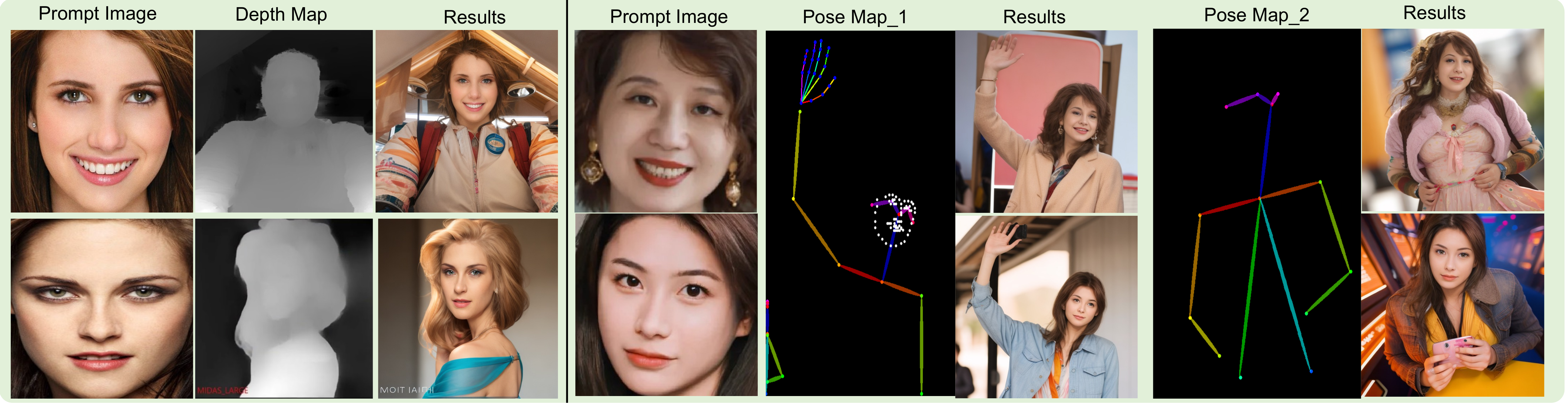}
    \caption{The proposed Inv-Adapter adapted to ControlNet~\cite{zhang2023adding} can get more diverse results.}
    \label{fige7}
    \vspace{-0.3cm}
\end{figure*}

\begin{figure*}[t!]
    \centering\setlength{\abovecaptionskip}{0.1cm}
    \includegraphics[width=0.99\linewidth]{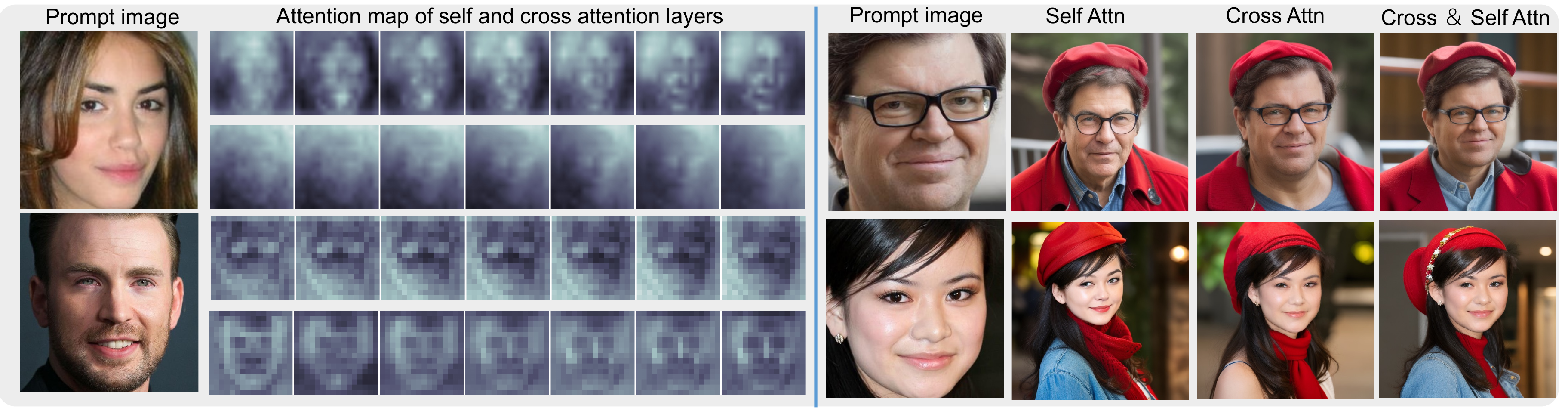}
    \caption{Left: attention maps of prompt images in the self attention and cross attention layers of the EAA. Right: ablation results of EAA adopted in the self attention layer only, the cross-attention layer only, and both, respectively.}
    \label{fig8}
    \vspace{-0.3cm}
\end{figure*}
\myparagraph{With ControlNet.}
To further demonstrate the superiority of Inv-Adapter, we demonstrate the diversity generation capability of the proposed method joint ControlNet~\cite{zhang2023adding}. As shown in Figure~\ref{fige7}, we show the generation results based on depth maps and pose maps, respectively, which also demonstrates that the proposed method can be effectively compatible with the community model (Supplementary Material provides the results of introduce other base models).


\myparagraph{Generation Speed and Trainable Parameters.}
To more comprehensively analyze the efficiency of the proposed method, we exhibit the comparison of the trainable parameters and generation speed of Inv-Adapter with recent state-of-the-art methods in Table~\ref{tab3}. 
It is clear that IP-Adapter~\cite{ye2023ip} yields smaller training parameters and high generation speed. However, its ID fidelity is poor.
InstantID~\cite{wang2024instantid} requires more training resources due to the IdentityNet.
Our method achieves a good balance in terms of generation speed and training parameters.

\subsection{Ablative Study}

\myparagraph{Embedded Attention Adapter.}
%
We verify the performance of Inv-Adapter's embedded attention adapter embedded only in the self or cross attention layers, respectively, and the results are shown in Figure~\ref{fig8} (quantitative results are presented in the \textbf{supplementary material}). 
A large number of generation results demonstrate that EAA is ineffective only in the cross attention layer or only in the self attention layer. However, when they work simultaneously, the results obtained are advanced.
Attention maps also indicate that the Inv-Adapter captures more details of the face.

\begin{figure*}[t]
\centering
\begin{minipage}[t]{0.64\textwidth}
\centering\setlength{\abovecaptionskip}{0.1cm}
\includegraphics[width=0.99\linewidth]{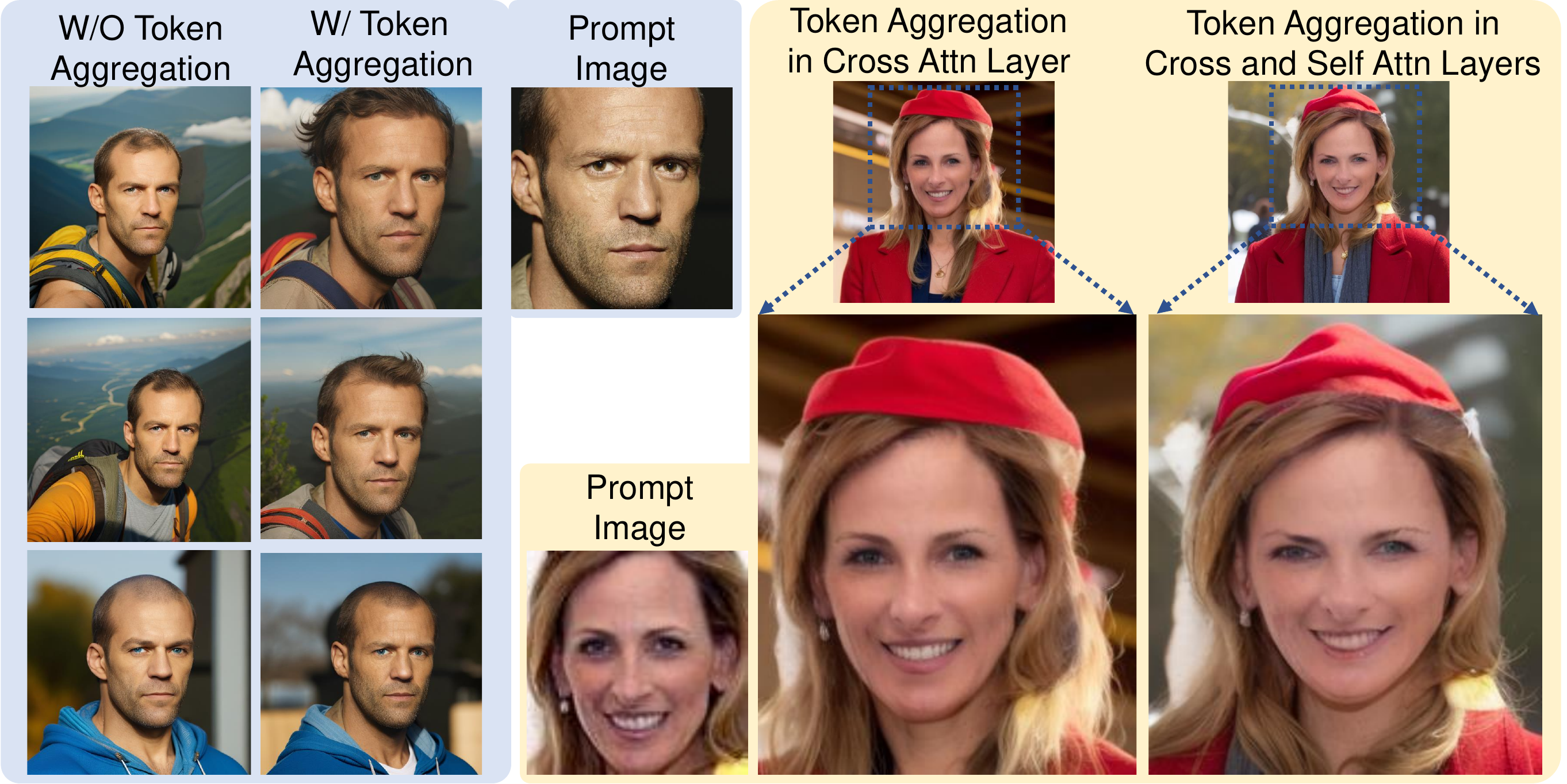}

	\caption{    %
    Ablation experiment results for token aggregation.
    }
	\label{fig9}
\end{minipage}
\begin{minipage}[t]{0.05\textwidth}
\end{minipage}
\begin{minipage}[t]{0.27\textwidth}
\centering
 \includegraphics[width=0.99\linewidth]{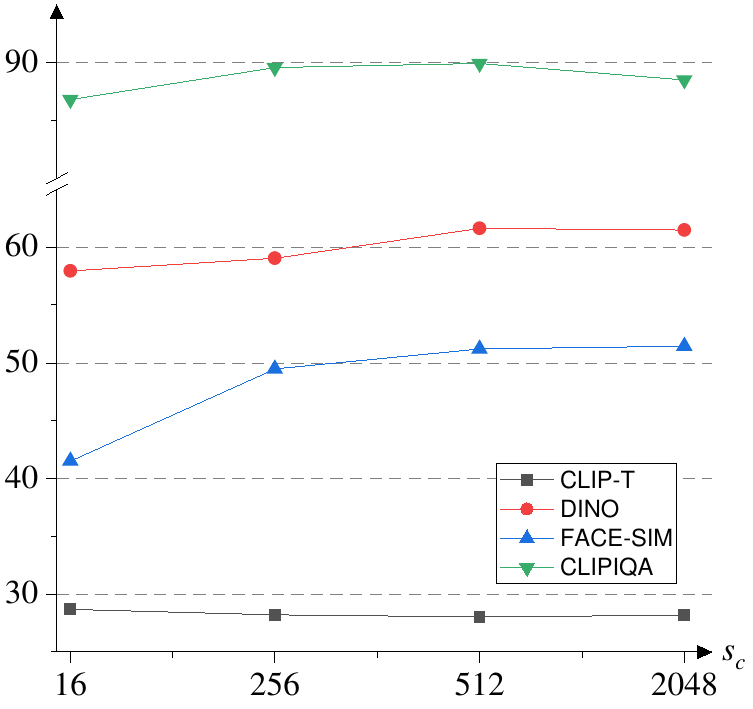}\setlength{\abovecaptionskip}{0.1cm}
    \caption{Ablation results of aggregation num $s_c$. }
    \label{fig10}

\end{minipage}
\vspace{-0.4cm}
\end{figure*}

\myparagraph{Token Aggregation.} The ablation experiments for token aggregation are shown on the left side of Figure~\ref{fig9}, which improves ID fidelity. 
However, we get a negative result when introducing token aggregation to the self-attention layer. 
As shown on the right side of Figure~\ref{fig9}, token aggregation in the self-attention layer leads to smearing and splicing in the face region of the generated images. 
As we know, self-attention layer focuses on the fusion of the intermediate features themselves, while the cross attention layer focuses on the interaction of the intermediate features with the text embeddings. The extracted diffusion features are essentially aligned with the intermediate features and thus can be directly embedded in the self attention layer. While in the cross attention layer, it cannot be injected directly similar to text features. Token aggregation can alleviate the modality gap, which facilitates the exploration of detailed information on ID diffusion features.
%

\myparagraph{Aggregation Number.}
Figure~\ref{fig10} illustrates the ablation experiment results for the aggregation number $s_c$.
It can be observed that as $s_c$ increases, the face ID fidelity (FACE-SIM) improves but the loyalty (CLIP-T) decreases. 
The possible reason is that as the increase in $s_c$, the diffusion and intermediate features are fused more effectively. Meanwhile, the number of tokens for diffusion features significantly exceeds text features, resulting in the overwhelming influence of the prompt image over the text prompt.
From Figure~\ref{fig10}, setting $s_c$ to 256 or 512 yields the best results.
%

\subsection{Limitation} \label{sec:4.4}
(1) Despite the impressive results achieved by Inv-Adapter, the current training data fails to fully explore the potential of such a novel framework. The trainging datasets of the current version generally look directly at the camera, lacking diversity to some extent. We believe this problem can be mitigated by involving ID-oriented data construction pipeline like PhotoMaker~\cite{li2023photomaker}. (2) Speed bottleneck. Despite the significant reduction in model size, image inversion reduces speed, and one possible solution is to use LCM model acceleration.

\section{Conclusion} \label{sec5}
\label{sec:Conclusions}
For the ID customization generation task, we propose Inv-Adapter based on image inversion and a lightweight adapter.
First, we use DDIM inversion to map the prompt image to the noise domain and extract the diffusion features of the face image during its denoising process, which is more aligned with the text-to-image model to explore more face details.
 Second, we propose an embedded attention adapter to efficiently inject ID diffusion embedding into the self and cross attention layers of the text-to-image model.
  Extensive results show that Inv-Adapter uses the \textit{smallest trainable parameters and model size} to obtain competitive results compared to recent advanced methods.
  %
{\small
\bibliographystyle{ieee_fullname}
\bibliography{egbib.bib}
}

\end{document}